\documentclass[11pt,a4paper]{article}
\usepackage[hyperref]{nlp4MusA}
\usepackage{times}
\usepackage{latexsym}
\usepackage{todonotes}
\usepackage{algorithm}
\usepackage{algorithmic}
\usepackage{devanagari}
\usepackage{url}

\aclfinalcopy 

\title{Bollyrics: Automatic Lyrics Generator for Romanised Hindi}

\author{Naman Jain, Ankush Chauhan, Atharva Chewale,\\ \textbf{Ojas Mithbavkar, Ujjaval Shah, Mayank Singh} \\
  Indian Institute of Technology Gandhinagar, Gujarat, India \\
  \texttt{singh.mayank@iitgn.ac.in} \\}

\begin{document}
\maketitle
\begin{abstract}
Song lyrics convey a meaningful story in a creative manner with complex rhythmic patterns. Researchers have been successful in generating and analyisng lyrics for poetry and songs in English and Chinese. But there are no works which explore the Hindi language datasets. Given the popularity of Hindi songs across the world and the ambiguous nature of romanized Hindi script, we propose \textit{Bollyrics}, an automatic lyric generator for romanized Hindi songs. We propose simple techniques to capture rhyming patterns before and during the model training process in Hindi language. The dataset and codes are available publicly at \url{https://github.com/lingo-iitgn/Bollyrics}.
\end{abstract}

\section{Introduction}
Hindi movie songs, popularly known as Bollywood songs, have played an integral part in shaping the Indian Subcontinent's music culture. From its emergence in \textit{Alam Ara} (1931), India's first sound movie, Bollywood music evolved into a mainstream music genre over the decades due to its occurrence in Indian cinema and the dance performances. Motivated by automating the generation of creative content and understanding the rhyming structure of Hindi songs, we propose \textit{Bollyrics}, a lyrics generation system based on Bollywood songs.

With the advent of deep learning in natural language processing, machines learn and predict the context with high precision. Researchers have developed highly advanced language models such as BERT \cite{devlin2018bert} and GPT-3 \cite{brown2020language} to generate human-like texts by training billions of parameters and text documents. These models have shown promising results in generating human-readable and semantically significant creative content. Hence, we approach our problem as a text generation task. Apart from utilizing the power of text generation models to capture context, \textit{Bollyrics} capture the song structure (i.e., account for the number of paragraphs and lines), and rhythmic patterns (E.g., AABB, ABBA, ABCD) to generate a particular song. Such properties distinguish normal text from lyrics/songs. Since majority of the work is conducted in English (\citet{potash2015ghostwriter}, \citet{peterson2019generating}) and Chinese languages (\citet{wu2017chinese}, \citet{yi2018chinese}) we have introduced an unique approach to account for rhyming patterns in Hindi language. 

\begin{figure}
    \centering
    \resizebox{\hsize}{!}{
    \includegraphics{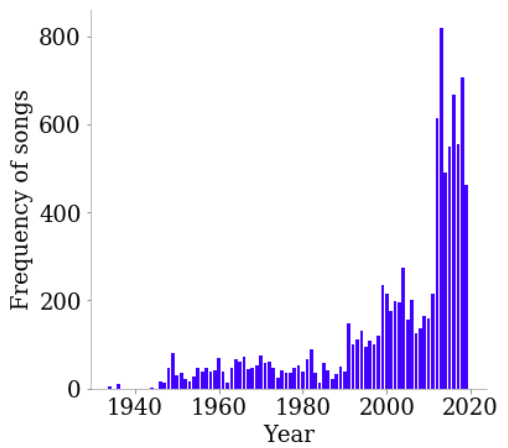}}
    \caption{Frequency distribution of songs released between 1934--2019. Majority of the songs are released in the $21^{st}$ century.}
    \label{fig:freq}
\end{figure}

\begin{table}
    \centering
    \begin{tabular}{c|c}
        \hline
        \textbf{Songs} & 10,229\\
        \textbf{Lines} & 366,219\\
        \textbf{Avg. lines (one song)} & 35.80\\
        \textbf{Tokens} & 2,094,427 \\
        \textbf{Unique tokens} & 71,135 \\
      \textbf{Year Range} & 1934 - 2019 \\
        \hline
    \end{tabular}
    \caption{Statistics of dataset used to train \textit{Bollyrics}}
    \label{tab:tab1}
\end{table}

\textit{Outline of the paper}: Section \ref{sec:data} introduces the dataset. In Section \ref{sec:method}, we present the pipeline for setting up, training and inferencing \textit{Bollyrics}.  We discuss the evaluation and results in Section \ref{sec:results}. Finally, we discuss the limitations of this paper in Section \ref{sec:limit} and conclude the paper in Section \ref{sec:Conclu}.

\begin{table*}
\begin{tabular}{|c|c|c|c|}
\hline
\multicolumn{2}{|c|}{Devanagri Sound} & \multicolumn{2}{c|}{Romanized word ending} \\ \hline
\multicolumn{2}{|c|}{{\dn e}} & \multicolumn{2}{c|}{\begin{tabular}[c]{@{}c@{}}-ein (e.g. humein), - e (e.g. tujhe), - ey (e.g. abey), \\ words- hai, hain\end{tabular}} \\ \hline

\multicolumn{2}{|c|}{{\dn a}/{\dn aA}} & \multicolumn{2}{c|}{\begin{tabular}[c]{@{}c@{}}-aa (e.g. bataa), -a (e.g. kaha), -yan (e.g. duriyan),\\ -yaan (e.g. yaariyaan), -uan (e.g. dhuan)\end{tabular}} \\ \hline

\multicolumn{2}{|c|}{{\dn u}/{\dn U}} & \multicolumn{2}{c|}{\begin{tabular}[c]{@{}c@{}}-u (e.g. kahu), -oo (e.g. bandhoo), -oon (e.g.\\ kahoon), -uun (e.g. jaauun), -un (e.g. bataaun)\end{tabular}} \\ \hline

\multicolumn{2}{|c|}{{\dn i}/{\dn I}} & \multicolumn{2}{c|}{\begin{tabular}[c]{@{}c@{}}-i (e.g. kahi (exceptions- hai, mai, jai), -ee (e.g.\\ chalee), -iin (e.g. nahiin), \\ -in (exceptions- hain, main)\end{tabular}} \\ \hline

\multicolumn{2}{|c|}{{\dn d}} & \multicolumn{2}{c|}{-d} \\ \hline

Devanagri Sound & Romanized word ending & Devanagri Sound & Romanized word ending \\ \hline
{\dn k} & -k & {\dn h} & -h \\ \hline
{\dn r} & -r & {\dn m} & -m \\ \hline
{\dn t} & -t & {\dn l} & -l \\ \hline
{\dn b} & -b & {\dn p} & -p \\ \hline
\end{tabular}
\caption{Mapping of Devanagri sounds with ending characters of romanized hindi }\label{fig:fig1}
\end{table*}
\section{Dataset}
\label{sec:data}
Since there is no open-source lyrical dataset in the Hindi language, we introduce a new dataset of Bollywood lyrics compiled using JioSaavn\footnote{jiosaavn.com} and LyricsBogie\footnote{lyricsbogie.com} services. We compiled the dataset after removing duplicates while merging (original songs and remix version with the same name), non-Hindi songs (E.g., songs in Punjabi, Bengali, Sanskrit hymns, Telugu, English and various Indian languages), non-Roman script, sound effects written in words and songs with incomplete metadata. Finally, the statistics of the dataset obtained are displayed in Table~\ref{tab:tab1}. Figure~\ref{fig:freq} shows the yearly frequency distribution of the song release from 1934 to 2019. 

\section{Methodology}
\label{sec:method}
\subsection{Setting up \textit{Bollyrics}}
\label{sec:method1}

To generate lyrics with rhyming schemes, we need to apply a grouping technique to collect words with similar ending sounds. As opposed to English, all the characters sound corresponds to a word pronunciation in Hindi. For e.g., ``Colonel'' is pronounced as ``k\raisebox{\depth}{\rotatebox{180}{e}}rnl'' by skipping the character ``l'' and adding new sounds in English, whereas in Hindi ``Colonel'' is written as ``karnal'', whose character's sounds aligns with it's pronunciation.  Hence, we group the words based on the last syllable in our corpus. Though it is trivial to group together words in the original script of a language, the machine-translated version of a language may result in varying ending letters. E.g., the Hindi word for ``we'' can be written as ``humein'', ``hume'', and ``humey'' in romanized version. 
Since our dataset is in the romanized script, we grouped the words by manually mapping a word's ending character with it's corresponding sound in Devanagri script to account for variation. We performed this process on top 1000 most-occurring ending words based on our understanding of Hindi langauge. The above process reduces the vocabulary to 70,731 from 71,153. Finally, we automatically mapped the ending character with its Devanagri sound to get 13 groups of sounds shown in Table \ref{fig:fig1}.

We separate all the songs into independent lines and extract the last two words from each line. Then, we allocate the pair of words into the Devanagri sound group \ref{fig:fig1} of the last word in the line. For instance, "Kaise Tumhe Roka Karun" is a line from a popular song named "Agar Tum Saath Ho" by A.R. Rahman. We extract the pair $\{$"roka", "kar\underline{un}"$\}$ and allocate it to the {\dn u}/{\dn U} Devanagri group.   
 
 We can now incorporate multiple existing text generation models in NLP to learn the corpus and generate meaningful lyrics. For this study's scope, we build \textit{Bollyrics} by training two language models: an N-Gram-based model and an LSTM-based \cite{hochreiter1997long} text generation model.
 \begin{itemize}
     \item \textbf{N-Gram-based model}: We learn the tri-gram probabilities using all the lines in the corpus to build an N-Gram text generator. The choice of tri-gram over bi-grams and quad-gram is due to the fact that either the meaning does not get captured accurately (bi-gram), or the model loses it's ability to generate creative sequences (quad-gram).
     \item \textbf{LSTM-based model}: We build a model with 128 units and two LSTM layers, which takes an embedding of size 50 generated from a sequence of length 25 and vocabulary of 70,731 words. On top of the two layers, we added a Dense layer of 128 neurons and ``Relu'' activation function. Our output layer is a ``Softmax'' layer of size is equivalent to the vocabulary i.e., 70,731. We trained the model using categorical cross-entropy, stochastic gradient descent with momentum 0.9, batch size of 256 and a learning rate of 0.001 for 1000 epochs on Google Cloud VM equiped with NVidia V100 GPU, 32GB RAM and 8 CPU cores. We initialized the weights with the orthogonal initializer.
 \end{itemize}
 
We trained both the models with a time step of two words (in case of LSTM, while tri-gram inherently consists of two time-steps) in reverse direction to ensure that the last word follows the rhyming scheme. For instance, the line ``Kaise Tumhe Roka Karun'' will be trained as follows: $\{$``karun'', ``roka''$\}$ $\rightarrow$ ``tumhe'', $\{$``roka'', ``tumhe''$\}$ $\rightarrow$ ``kaise'' and $\{$``tumhe'', ``kaise''$\}$ $\rightarrow$ ``$<$start$>$'', here the arrow connects the input(left) and ground truth(right).

\subsection{Generating lyrics from \textit{Bollyrics}}
\label{sec:method2}

To generate lyrics through \textit{Bollyrics}, the user can provide a rhyme scheme for the paragraph. Rhyme scheme can be of any form - AABB/ABAB/AAA/ABCD and many more. We randomly allocate a unique group from Table \ref{fig:fig1} to each letter in the scheme without replacement. The number of letters in the scheme denotes the number of lines in each paragraph, and the number of paragraphs in the generated song is decided by the user. Then, we start iterating through each letter in the scheme and select a pair of words randomly from the group corresponding to that letter. We reverse the order of the words and inference the model. Using the outputs and a time step of two words, we sequentially generate a sentence until '$<$start$>$' token appears or the sentence length becomes more than 10. Then, we reverse the generated sentence's ordering to obtain a line of the lyrics. We repeat the steps for all the paragraphs. Figures  \ref{fig:my_label} and \ref{fig:example2} are the examples of the output generated from \textit{Bollyrics} for ABAB and ABBA rhyme schemes respectively.


\begin{figure}[t]
\centering
\small
\begin{tabular}{l}
\begin{tabular}{l}
{\color[HTML]{333333} \begin{tabular}[c]{@{}l@{}} $<$start$>$ yun aku ta ruswaii mujhako \textbf{sharmaata \textcolor{red}{tha}}\\$<$start$>$ beete saaye se hone \textbf{laga \textcolor{blue}{hoon} }\\ $<$start$>$ kabhi \textbf{sochataa  \textcolor{blue}{hoon} }\\ $<$start$>$ main khel kahu \textbf{ya \textcolor{red}{nindiya} }\end{tabular}} \\ \\
\begin{tabular}[c]{@{}l@{}}$<$start$>$ mamta ki bahoon \textbf{mein \textcolor{red}{khudaa} }\\$<$start$>$ meri dil ki sang na \textbf{rang \textcolor{blue}{churaau}}\\$<$start$>$ ye yad dekha yeh to mai \textbf{kya \textcolor{blue}{karu}}\\$<$start$>$ tere woh socha mein wo keh \textbf{laage \textcolor{red}{jara}}\end{tabular} \\ \\
\begin{tabular}[c]{@{}l@{}}$<$start$>$ duniya hu dil se tu mujhe kya \textbf{kho \textcolor{red}{jayega}} \\$<$start$>$ socha ko baatein kar \textbf{raha \textcolor{blue}{hoon} }\\$<$start$>$ teri main \textbf{ehsaas \textcolor{blue}{hoon} }\\$<$start$>$ ye dil tera \textbf{ho \textcolor{red}{gaya} }\end{tabular} \\ \\
\begin{tabular}[c]{@{}l@{}} $<$start$>$ fir waapas meri arre \textbf{hai \textcolor{red}{vidhaataa} }\\$<$start$>$ jo haath kaam ud khul main \textbf{sharamaati \textcolor{blue}{phiru}}\\$<$start$>$ phir izaajat khud main \textbf{bata \textcolor{blue}{doon} }\\$<$start$>$ aisa tune khuda \textbf{gagan \textcolor{red}{chala}}\end{tabular}
\end{tabular}
\end{tabular}
    \caption{Example of \textcolor{red}{A}\textcolor{blue}{B}\textcolor{blue}{B}\textcolor{red}{A} rhyme scheme (LSTM model). \textbf{Bold} words are the pair of words selected from the rhyming buckets: \textcolor{red}{{\dn a}/{\dn aA}} and \textcolor{blue}{{\dn u}/{\dn U}}}
    \label{fig:example2}
\end{figure}

\begin{figure}[t]
\centering
\small
\begin{tabular}{l}
\begin{tabular}[c]{@{}l@{}}$<$start$>$ kya hai mere \textbf{liye \textcolor{red}{jindagi}}\\ $<$start$>$ ayi mein main \textbf{khona \textcolor{blue}{chaahoon}}\\ $<$start$>$ zulfon ki socho uthi \textbf{yad \textcolor{red}{aayi}}\\ $<$start$>$ tumhein main \textbf{kyaa \textcolor{blue}{bataaoon}}\end{tabular} \\ \\
\begin{tabular}[c]{@{}l@{}} $<$start$>$ kis naheen \textbf{jaa \textcolor{red}{hansi}}\\ $<$start$>$ palkein jhukaun tujhe dil \textbf{mein \textcolor{blue}{basaau}}\\ $<$start$>$ koi chhavi \textbf{to \textcolor{red}{hogi}}\\ $<$start$>$ jab mere ghar se main \textbf{bataa \textcolor{blue}{doon}}\end{tabular} \\ \\
\begin{tabular}[c]{@{}l@{}} $<$start$>$ hey mat kaali \textbf{neend \textcolor{red}{churaai}}\\$<$start$>$ jise main khud kaa behka \textbf{aayi \textcolor{blue}{hu}}\\ $<$start$>$ jaise taron kii baat sune \textbf{raat \textcolor{red}{suhaani}}\\ $<$start$>$ main teri\textbf{ kitaab \textcolor{blue}{hoon}}\end{tabular} \\ \\
\begin{tabular}[c]{@{}l@{}}$<$start$>$ hothon pe naa karti meri\textbf{ ladki \textcolor{red}{razi}}\\ $<$start$>$ main \textbf{happy \textcolor{blue}{hoon}}\\ $<$start$>$ kesi tu jo \textbf{chaa \textcolor{red}{gayi}}\\ $<$start$>$ to kisi ke harfedua \textbf{main \textcolor{blue}{hoon}}\end{tabular}
\end{tabular}
    \caption{Example of \textcolor{red}{A}\textcolor{blue}{B}\textcolor{red}{A}\textcolor{blue}{B} rhyme scheme (LSTM model). \textbf{Bold} words are the pair of words selected from the rhyming buckets: \textcolor{red}{{\dn i}/{\dn I}} and \textcolor{blue}{{\dn u}/{\dn U}}}
    \label{fig:my_label}
\end{figure}

\section{Result and Discussion}
\label{sec:results}

To evaluate the songs, we conducted a study by asking humans to tell whether the lyrics make sense or not. We asked each human annotator to label 50 songs generated by the each of the models in Section \ref{sec:method1}, with varying rhyming schemes. The annotators had to label each paragraph as ``Makes sense'', if the paragraph follows a rhyming scheme, the sentences are meaningful, and the context of the paragraph revolves around a theme else the annotator may label the paragraph as ``Does not make sense''. Since our model outputs songs with four paragraphs, each annotator had to label 200 paragraphs. 

\begin{figure}
    \centering
    \resizebox{\hsize}{!}{
    \includegraphics{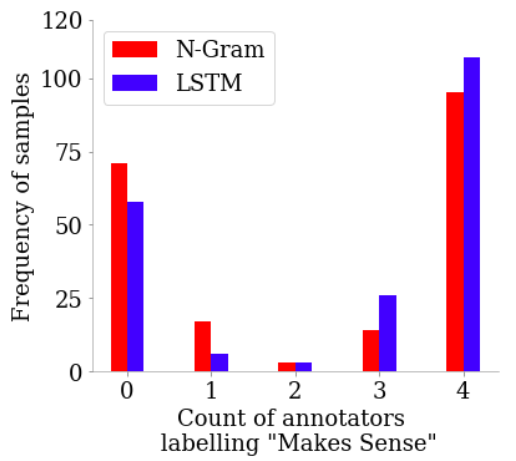}}
    \caption{Human evaluation of \textit{Bollyrics}}
    \label{fig:annot}
\end{figure}

We chose four Hindi-speaking annotators for the study. We combined the results and calculated Kappa score \cite{fleiss1971measuring} to fulfill the acceptance of human collected data by inter-annotator agreement for many raters. For Kappa Score, N=200, k=2, and n=4. We calculated the Kappa scores separately for LSTM and N-Gram generated texts. The scores came out to be 0.82 (LSTM) and 0.80 (N-Gram). We can consider this score to be in correspondence with near-complete agreement \cite{dhar2018enabling}. We, further, theorize that such a high agreement could arise because the annotators have similar backgrounds and similar perception of Bollywood music. The detailed results of the study are displayed in Figure \ref{fig:annot}. Overall, 54.5\% of the paragraphs in N-Gram based model outputs and 64.5\% of the paragraphs in LSTM based model outputs were labeled as ``Makes Sense''. We calculate this metric after considering that the ``Makes Sense'' class contains samples that are labeled as ``Makes Sense'' by more than two annotators.

\section{Limitations}
\label{sec:limit}

First limitation is that we did not accounted for the variation (mentioned in Section \ref{sec:method1}) in the whole dataset. The complexity and the extent of variation in romanized Hindi can be seen by the example of the sentence -``ye jaruri hai'', which translates to ``It is necessary'' in English. We can write the sentence as ``ye(h) (j/z)ar(u/oo)r(i/ee) h(a/e)(i/in)'' - which gives us 64 ways to write the same sentence in romanized Hindi. This variation leads to an unwanted increase in vocabulary by adding multiple forms of the same word. This phenomenon will give rise to a skewed distribution of probability mass to multiple forms, thereby adding noise to the training process. Also, while allocating groups automatically the words may even end up in the wrong sound group of Table \ref{fig:fig1}.

Second limitation while working with the romanized Hindi dataset is that some words in romanized Hindi have a meaning in the English language. For instance, ``main'' in Hindi means ``Myself'' but ``main'' is also an English word which means ``chief in size or importance''. This ambiguity may add up confusion for the machine in terms of the context.

\section{Conclusion and Future Work}
\label{sec:Conclu}
In this paper, we have introduced a lyrics generation system that produces creative lyrics that make sense for a human being. In the future, we may work on normalizing the word variations mentioned in Section \ref{sec:limit}. Further, one can experiment with more complex architectures such as GANs, seq2seq, and GPT-3. Instead of structuring the input, we can manipulate the architecture into marking the rhyme scheme by itself. We might also make the architecture learn the mood, emotion, and the lyricist style associated with the song.

\bibliography{nlp4MusA}
\bibliographystyle{nlp4MusA_natbib}

\end{document}